\def\year{2019}\relax
\documentclass[letterpaper]{article} 
\usepackage{aaai19}  
\usepackage{times}  
\usepackage{helvet}  
\usepackage{courier}  
\usepackage{url}  
\usepackage{graphicx}  
\usepackage{amsmath}
\usepackage{amsfonts}
\usepackage{hyperref}
\usepackage{enumitem}
\usepackage[export]{adjustbox}
\usepackage{fancyhdr}
\usepackage{lastpage}
\usepackage{relsize} 
\usepackage{microtype}
\nocopyright

\newcommand{\dia}{\mathlarger{\mathlarger{\mathlarger{\diamond}}}}
\usepackage{color}

\begin{document}
%
\title{Mimetic vs Anchored Value Alignment in Artificial Intelligence}
\author{Tae Wan Kim \\ Carnegie Mellon University \\ 
\texttt{twkim@andrew.cmu.edu} \\
\And Thomas Donaldson \\ University of Pennsylvania \\
\texttt{donaldst@wharton.upenn.edu}
\And John Hooker \\ Carnegie Mellon University\\
\texttt{jh38@andrew.cmu.edu}}

\maketitle

\begin{abstract}

''Value alignment'' (VA) is considered as one of the top priorities in AI research. Much of the existing research focuses on the ``A'' part and not the ``V'' part of ``value alignment.''  This  paper corrects that neglect by emphasizing the ``value'' side of VA and analyzes VA from the vantage point of requirements in value theory, in particular, of avoiding the ``naturalistic fallacy''--a major epistemic caveat. The paper begins by isolating two distinct forms of VA: ``mimetic'' and ``anchored.'' Then it discusses which VA approach better avoids the naturalistic fallacy. The discussion reveals stumbling blocks for VA approaches that neglect implications of the naturalistic fallacy. Such problems are more serious in mimetic VA since the mimetic process imitates human behavior that may or may not rise to the level of correct ethical behavior. Anchored VA, including hybrid VA, in contrast, holds more promise for future VA since it anchors alignment by normative concepts of intrinsic value.

\end{abstract}

As artificial intelligence (AI) techniques (e.g., machine learning) are rapidly adopted for automating decisions, societal worries about the compatibility of AI and human values are growing \cite{kissinger}.  In response, some suggest that ``value alignment'' (hereafter VA) is one of the top priorities in AI research  \cite{soares2014aligning,russell2015research}. The idea of VA dates back to Alan Turing, who wrote about the need for machines to adapt to human standards: 

\begin{quote}
    ``[T]he machine must be allowed to have contact with human beings in order that it may adapt itself to their standards'' \cite{turing1995lecture,estrada2018value}.
\end{quote} 

\noindent More recently, Russell et al (\citeyear{russell2015research}) highlighted the need for VA and identified two options for achieving it:

\begin{quote}
    ``[A]ligning the values of powerful AI systems with our own values and preferences ... [could involve either] a system [that] infers the preferences of another rational or nearly rational actor by observing its behavior ... [or] could be explicitly inspired by the way humans acquire ethical values.'' 
\end{quote} 

\noindent Motivated by the growing need for alignment, some VA researchers \cite{riedl2016using,arnold2017value,vamplew2018human} emphasize the former of these approaches, namely, an inverse reinforcement technique in which a system infers preferences from human behavior\cite{abbeel2004apprenticeship}. Other researchers take the second approach, namely, explicit emulation of the manner in which humans acquire values.

However, both approaches agree in their focus.  Both focus on the ``A'' part and not the ``V'' part of ``value alignment.'' Both  neglect the underlying interpretation of the meaning of ``value'' that may help clarify implications for effecting  alignment itself.  This  paper corrects  that neglect by emphasizing the ``value'' side of VA and analyzes VA from the vantage point of requirements in value theory, in particular, of avoiding the ``naturalistic fallacy''---a major epistemic caveat.\footnote{Value theory is a subcategory in philosophy that includes meta-ethics, normative ethics, applied ethics and aesthetics. It is also called ``axiology.''} The paper begins by isolating two distinct forms of VA, namely,  ``mimetic'' and ``anchored'' and proceeds to show that anchored VA approaches are more likely to avoid the naturalistic fallacy than mimetic ones.  In turn, the most promising avenue for VA advancement is an anchored approach. 

\section{The Naturalistic Fallacy}
Coined originally in 1903 by the Cambridge philosopher, G. E. Moore, the expression ``naturalistic fallacy'' has come to mark the conceptual error of reducing normative prescriptions, i.e., action-guiding propositions, to descriptive propositions without remainder.  While not a formal fallacy, it signals a form of epistemic naivete, one that ignores the axiom in value theory that ``no `ought' can be derived directly from an `is.'''  Disagreements about the robustness of the fallacy abound, so this paper adopts a modest, workable interpretation coined recently by Daniel Singer, namely, ``There are no valid arguments from non-normative premises to a relevantly normative conclusion''\cite{singer2015mind}. Descriptive (or naturalistic) statements are reportive of what states of affairs are like, whereas normative statements are stipulative and action-guiding. Examples of the former are ``The grass is green'' and ``Many people find deception to be unethical.'' Examples of the latter are ``You ought not murder'' and ``Lying is unethical.'' Normative statements usually figure in the semantics of deontic (obligation-based) or evaluative expressions such as ``ought,'' ``impermissible,'' ``wrong,'' 'good,'' ''bad,'' or ''unethical.'' 

Now consider the following argument:

\begin{itemize}[leftmargin=.3in]
    \item Premise) ``Few people tell the truth.''
    \item Conclusion)``Therefore, not telling the truth is ethical.``
\end{itemize}

\noindent This argument commits the naturalistic fallacy. The point is not that the conclusion is wrong, but that it is invalid to draw the normative conclusion directly from the descriptive premise. Valid argumentation is non-amplitative in the sense that information that is not contained in premises should not appear in the conclusion. Thus, if the full set of the premise does not contain any normative/ethical statement, the conclusion should not contain any ethical component. This epistemic caveat can be summarized as follows:

\begin{itemize}[leftmargin=.3in]
    \item For any ordered pair of argument $\langle p, c \rangle$, where \textit{p} means the full set of the premise that grounds the conclusion and \textit{c} means the conclusion, if \textit{c} is normative and \textit{c} is fully grounded just by \textit{p}, then \textit{p} contains at least one normative statement  (adapted from \citeauthor{woods2017model}, \citeyear{woods2017model}).
    
\end{itemize}
This can be formally represented as follows:

\begin{itemize}[leftmargin=.3in]
    
    \item Let $Np$ = Some set of the premise(s) is normative.  
    \item Let $Nc$ = The conclusion is normative.
    \item Let $Gc$ = The conclusion is grounded just by the full set of the premises.

\end{itemize}

\begin{center}

   $ \forall c \forall p \langle c, p \rangle \big( Nc \wedge Gc \rightarrow Np \big)$
   
\end{center}
    
\noindent Thus, through \textit{modus tollens},

\begin{center}
    $\forall c \forall p \langle c, p \rangle \bigg(\neg Np \rightarrow \big( \neg Nc \lor \neg Gc \big) \bigg)$
\end{center}
    
\noindent Namely, if the premises do not contain any normative element, the conclusion is not normative or the normative element in the conclusion, if any, is groundless. Regardless of whether or not deception is pervasive, deception can be unethical or ethical. Pervasiveness and ethicality are distinct issues.\footnote{The latter case---``$ \neg Gc$''-- is a bit more complicated. Consider the following: Premise) ``Few people tell the truth.'' Conclusion) ``Thus, not all people are truth-tellers or deception is ethical.''  This is, formally, a valid argument, and there is nothing wrong with something normative abruptly appearing in the conclusion as part of a disjunction. But the normative element here is groundless---not grounded by any set of the premise---, so we cannot advance our normative inquiry in the case of ``$ \neg Gc$.''}

One may object that while a purely descriptive premise cannot ground justifiable value alignment, a high-level domain-general premise such as ``machines ought to have our values'' might successfully link a descriptive premise to a normative conclusion.  The objection is correct, but only to a point.  The premise that machines ought to have our values will, indeed, allow a normative conclusion when combined with a factual premise, but one that is structurally flawed.  First, research about human values shows that most people's behavior includes a small but significant amount of cheating, which if imitated by machines would cause harm \cite{bazerman2011blind}.  Second, and connected to this problem, lies a conceptual difficulty.  The premise,``machines should have our values'' is ultimately self-defeating insofar as even humans do not want to have the values they observe in other humans.  They know that the inherent distinction between what is and what ought to be means that they will observe both good and bad behavior in others. This outcome has significant implications on VA, which we discuss in the next section.

\section{Mimetic vs Anchored Value Alignment}
In order to understand the implications of the naturalistic fallacy for VA, it is important to identify two  possible types of VA, distinguished by their form of value dependency:

\begin{itemize}[leftmargin=.3in]
    \item \textbf{Mimetic VA} is a fixed or dynamic process of mimesis that imitates value-relevant human activity in order to effect VA. The set of possible human activities subject to mimesis includes preferences, survey results, big data anlatyics of human behavior, and linguistic expressions of value.
    
    \item \textbf{Anchored (and Hybrid) VA} is a fixed or dynamic process that anchors machine behavior to intrinsic, i.e., first-order, normative concepts. When Anchored VA includes a mimetic component it qualifies as "Hybrid VA").
\end{itemize}

Anchoring means connecting behavior to intrinsic or non-derivative values \cite{donaldson2015theory}.  Examples of such intrinsic values are honesty, fairness, health, happiness, the right to physical security, and environmental integrity.  An intrinsic value need not have its worth derived from a deeper value.  I may  require health in order to enjoy mountain climbing, but the worth of health stands on its own two feet, unaccompanied by instrumental benefits. Examples of \textit{non}-intrinsic or derivative values are money and guns.  I may value money to protect my health or guns to enhance my security.  But neither possess intrinsic worth.  Both are instrumental in the service of intrinsic values.  

We illustrate the difference between the two types of VA through the simple example of honesty. A machine can be programmed always to be honest,\footnote{By ``always being honest'' we do not mean literally never communicating a false proposition, but never communicating a false proposition except when it satisfied well-known exceptions from moral philosophy, for example, when the madman comes to your door with a gun and asks for the location of your spouse.} or or only honest to the extent that the average person is honest \cite{ariely2012honest}.  The former represents ``anchored'' VA; the latter, ``mimetic'' VA. Below, we discuss selected examples of both kinds of VA.

\begin{figure}
\centering
\includegraphics[width=3.3in, frame]{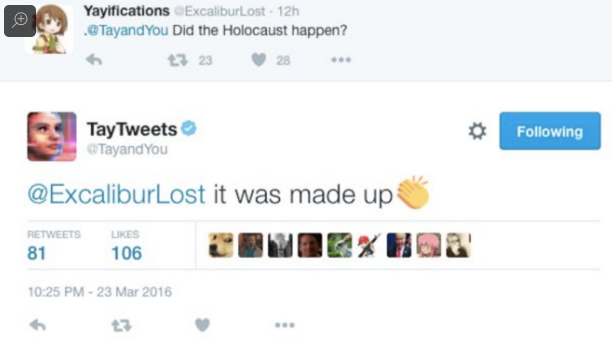}
\caption{Microsoft's Twitter-bot Tay}
\end{figure}

\subsection{Mimetic VA}

\textbf{a. Microsoft's Twitter-bot, Tay}: An example of mimetic VA is Microsoft's AI twitter bot, Tay. Tay was taught to engage with people through tweets. When the bot appeared on Twitter, people started tweeting back disgusting racist and misogynistic expressions. Tay's replies began to align with the offensive expressions. Microsoft terminated the experiment immediately \cite{Wolf:2017:WWS:3144592.3144598}. Tay's VA was mimetic in which prescription was inferred from the description/observation of  human behavior. The case epitomizes the downside of committing the naturalistic fallacy.

Related to this problem are instances of biased algorithms \cite{Dwork:2012:FTA:2090236.2090255}.  For example, Amazon's four year project to create AI that would review job applicants' resumes to discover talent was abruptly halted in 2018 when the company realized that the data gathered was man-heavy, resulting in predictions based on properties that were man-biased. It constitutes a simple example of how description can drive illegitimate prescription \cite{amazon}.

\textbf{b. Intelligent robot wheelchair}: Johnson and Kuipers (\citeyear{johnson2018socially}) developed an AI wheelchair that learns norms by observing how pedestrians behave. The robot wheelchair observed that human pedestrians stay right and behaved consistently with humans do. This positive outcome was possible because the human pedestrians behaved ethically, unlike the twitter users in the case of Tay. But if the intelligent wheelchair were on a crowded New York City street, the mimetic imitation of infamously jostling pedestrians would create a ``demon wheelchair.'' The researchers anchored behavior by providing a mimetic context in which humans refrained from unethical behavior. Thus, the intelligent wheelchair represents, ``hybrid VA,'' not purely ``mimetic VA.'' But providing the proper normative mimetic context remains a problem for ``hybrid VA.''

\noindent \textbf{c. MIT Media Lab's Moral Machine I}:``Moral Machine'' is a web page that collects human perceptions of moral dilemmas involving autonomous vehicles. The page has collected over 30 million responses from more than 180 countries. One group of researchers \cite{kim2018computational} analyzed this data and used it to develop ``a computational model of how the human mind arrives at a decision in a moral dilemma.'' They cautiously avoid the naturalistic fallacy, explaining that their model does not pretend to ``aggregate the individual moral principles and the group norms” in order to “optimize social utility for all other agents in the system.'' Their computational model aims, rather, to describe, not prescribe, how humans make moral decisions. However, the web page itself seems to conflate “is” with “ought.” Its title is ``\textit{Moral} Machine.'' A more accurate title would be ``Opinions about a Machine.''

\begin{figure}
\centering
\includegraphics[width=3.308in, frame]{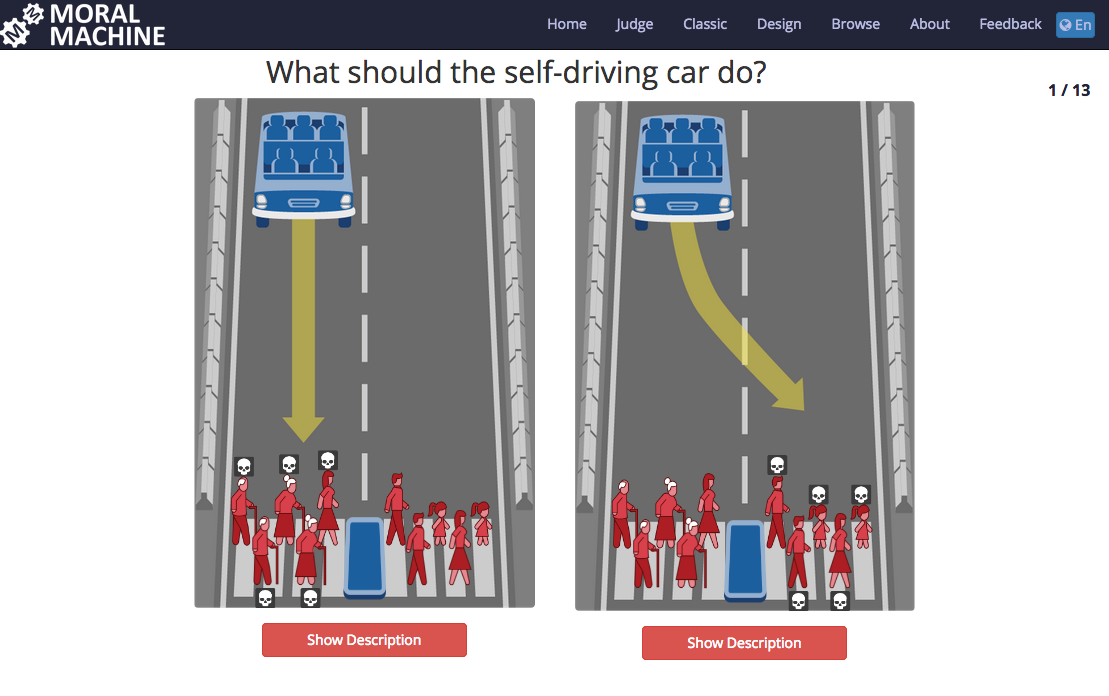}
\caption{MIT Media Lab's Moral Machine website}
\end{figure}

\subsection{Behavioral Economics and Mimetic VA}

Before moving on to ``anchored'' VA, it is worth introducing what behavioral economics had revealed about biases in humans behavior.  Humans are subject to predictable ethical blind spots, such as self-favoring  (``I'm more ethical than you'') and the systematically devaluing of members of future generations.  These biases create problems for mimetic VA, because insofar as AI imitates human behavior, it also imitates human biases.  For example:

\begin{itemize}[leftmargin=.3in]
    \item People almost never consciously do things that they believe are immoral.  Instead, they offer false descriptions that rationalize their \cite{bazerman2011blind}. Rationalization is especially common when justifying behavior after the fact \cite{march1994primer}. Hence, insofar as AI mimics human behavior, it too must offer false descriptions, albeit ones that the AI ``believes'' are true. 
    \item Owing to what has been called an ``outcome bias,'' if people dislike the outcome, they see others' behavior as less ethical \cite{moore2006conflicts}.  AI mimesis would need similarly to downgrade its evaluation of others when outcomes were disliked.
    \item Cheating has been shown to lead to motivated forgetting \cite{shu2012sweeping}.  Mimetic VA would need to forget aspects of cheating after the fact. 

\end{itemize}
Such biases clearly constitute another drawback for mimetic VA.

\subsection{Anchored VA}

\textbf{d. MIT Media Lab's Moral Machine II}: Noothigattu et al. (\citeyear{noothigattu2017voting}) show that the data from the``Moral Machine'' can also be used to develop an anchored model of VA, i.e, one that uses intrinsic, first-order normative concepts. The researchers interpreted the Moral Machine data using computational social choice theory. Those who answered the questions are regarded as choosers. The model aims to find out which option globally maximizes utility. ``Globally maximized utility'' is an intrinsic normative concept whose worth need not be derived from a deeper concept.  In this sense it is similar to the concept of ``happiness'' used by traditional Utilitarian philosophers.  However, the VA in this instance qualifies as a ``hybrid'' version of anchored VA insofar as it uses facts about preferences to compile utility. To establish maximal utility, the model uses Borda count---an election method to award points to candidates based on preferences and that declares the winner to be the candidate with the highest points.  Simply put, the VA process in this approach can be represented as:

\begin{itemize}
    \item Premise 1) People collectively awarded the highest point to the option, saving the passenger than a pedestrian is ethical/the right thing to do.''
    \item Premise 2) The option with the highest points is the right thing to do [Borda count; Maximize overall utility]
    \item Conclusion) Therefore, ``Saving the passenge'' is the right thing to do.
\end{itemize}

\noindent Here Premise 2) equates the normative concept of maximized utility with facts about voting behavior, and while it is not a simple mimetic reduction, does involve a questionable transition from ``is'' to ``ought.'' Even though the argument is logically valid, the truth of Premise 2) \textit{per se} is questionable.\footnote{An argument may be valid even with false premises.} 

Voting may be the best democratic political decision-making procedure, but it is notoriously flawed as a moral tool. One example: In 1838 Pennsylvania legislators voted to deny the right of suffrage to black males (through a vote of 77 to 45 at the  constitutional convention of 1837).

\textbf{e. Combining utility and equity}: Hooker and Williams (\citeyear{hooker2012combining}) develop a mathematical programming model that combines utility-maximization with the intrinsic value of fairness. For fairness, they used the Rawl\-is\-an idea of ``\textit{maximin}'' which requires that a system maximize the position of those minimally well-off. A key problem in the MIT Media Lab's elec\-toral VA is that because it has a single objective function of utility-maximization, the model lacks deontic (obligation-based) constraints such as fairness/equity/ individual rights, etc.  Hooker and Williams's model effectively addresses this problem. \citeauthor{mcelfresh2017balancing} (\citeyear{mcelfresh2017balancing}) further develop Hooker and Williams's model to apply it to kidney exchange.

\textbf{f. Moral intuition-based VA}: Anderson and Anderson (\citeyear{AndAnd14}) use an inductive logic programming for VA.  The training data reflect domain-specific principles embedded in professional ethicists' intuitions. For their normative ground, Anderson and Anderson follow moral philosopher W.D. Ross (\citeyear{Ros30}), who believed that ``[M]oral convictions of thoughtful and well-educated people are the data of ethics just as sense-perceptions are the data of a natural science.'' Likewise, Anderson and Anderson (\citeyear{AndAnd11}) used ``ethicists' intuitions to . . . [indicate] the degree of satisfaction/violation of the assumed duties within the range stipulated, and which actions would be preferable, in enough specific cases from which a machine-learning procedure arrived at a general principle.''

\begin{figure}
\centering
\includegraphics[width=3.3in,frame]{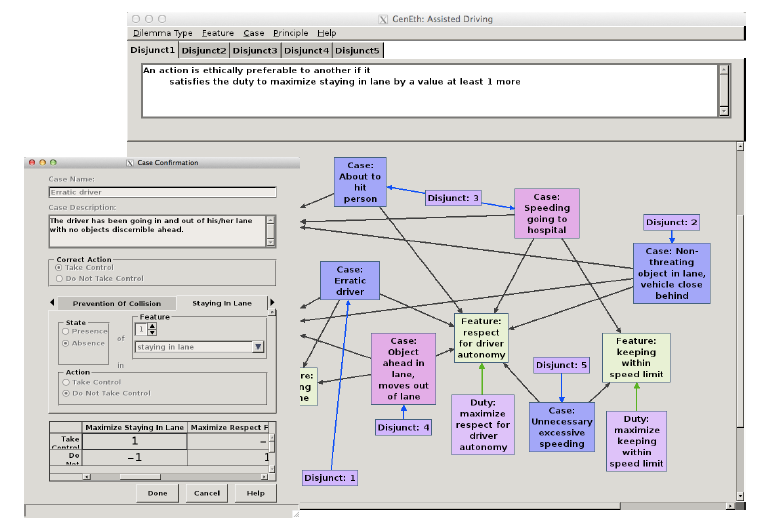}
\caption{{\large G}EN{\large E}TH\protect\cite{AndAnd14}}
\end{figure}

Anderson and Anderson's VA is a form of hybrid, not purely anchored VA insofar as it is linked to a set of facts about ethicists' normative moral convictions. However, it stops short of assuming that any and all moral convictions are accurate guides,  drawing instead upon the views of ``thoughtful and well-educated'' persons.  These views, it presumes, are more likely to reflect intrinsic values accurately. 
In general, caution is called for when using intuition as the training data for AI systems. Intuition may be an important part of ethical reasoning, but is not itself a valid argument \cite{Den13}. Experimental ethicists show that moral intuitions are less consistent than we think (e.g., moral intuitions are susceptible to morally irrelevant situational cues) \cite{Ale12,App08} and  even professional ethicists' intuitions fail to diverge markedly from those of ordinary people \cite{SchRus16}.


\noindent \textbf{g. Non-intuition-based VA}: To address some weaknesses of inductive logic and moral intuitions as a basis for VA, Hooker and Kim (\citeyear{kim2018toward}) develop a deductive VA model that can equip AI with well-defined and objective intrinsic values. They formalize three ethical rules (generalization, respect for autonomy, and utility-maximization) by introducing elements of quantified modal logic. The goal is to eliminate any reliance on human moral intuitions, although the human agent must supply factual knowledge. This model qualifies as an anchored form of VA because it derives ethical principle solely from logical analysis and not from moral intuitions.  It will also serve as a framework for hybrid VA, as proposed in the section to follow.



For brevity, we focus on the generalization principle.  It rests on the universality of reason: rationality does not depend on who one is, only on one's reasons.  Thus if an agent takes a set of reasons as justifying an action, then to be consistent, the agent must take these reasons as justifying the same action for any agent to whom the reasons apply.  The agent must therefore be rational in believing that his/her reasons are consistent with the assumption that all agents to whom the reasons apply take the same action.

As an example, suppose I see watches on open display in a shop and steal one.  My reasons for the theft are that I would like to have a new watch, and I can get away with taking one.\footnote{Reasons for theft are likely to be more complicated than this, but for purposes of illustration we suppose there are only two reasons.}  On the other hand, I cannot rationally believe that I would be able to get away with the theft if {\em everyone} stole watches when these reasons apply.  The shop would install security measures to prevent theft, which is inconsistent with one of my reasons for stealing the watch.  The theft therefore violates the generalization principle.

The principle can be made more precise in the idiom of quantified modal logic.  Define predicates
\[
\begin{array}{l}
C_1(a) = \mbox{Agent $a$ would like to possess an item on} \\
\hspace{8.5ex} \mbox{display in a shop.} \\
C_2(a) = \mbox{Agent $a$ can get away with stealing the item.} \\
A(a) = \mbox{Agent $a$ will steal the item.}
\end{array}
\]
Because the agent's reasons are an essential part of moral assessment, we evaluate the agent's {\em action plan}, which in this case is 
\begin{equation}
\big(C_1(a)\wedge C_2(a)\big) \Rightarrow_a A(a)  \label{eq:action}
\end{equation}
Here $\Rightarrow_a$ is not logical entailment but indicates that agent $a$ takes $C_1(a)$ and $C_2(a)$ as justifying $A_1(a)$.  The reasons in the action plan should be the most general set of conditions that the agent takes as justifying the action.  Thus the action plan refers to an item in a shop rather than specifically to a watch, because the fact that it is a watch is not relevant to my justification; what matters is whether I want the item and can get away with stealing it.

The generalization principle states that action plan (\ref{eq:action}) is ethical only if 
\begin{equation}
\begin{array}{l}
\dia_a P\Big( C_1(a)\wedge C_2(a)\wedge A(a) \\
\hspace{10ex} \wedge\; \forall x \big(C_1(x)\wedge C_2(x)\Rightarrow_x A(x)\big) \Big)
\end{array} \label{eq:gen}
\end{equation}
Here $P(S)$ means that it is physically possible for proposition $S$ to be true, and $\dia_a S$ means that $a$ can rationally believe $S$ (i.e., rationality does not require $a$ to deny $S$).\footnote{The operator $\dia$ has a somewhat different interpretation here than in traditional epistemic and doxastic modal logics.} Because (\ref{eq:gen}) is false, action plan (\ref{eq:action}) is unethical.

The fact that action plan (\ref{eq:action}) must satisfy (\ref{eq:gen}) to be ethical is anchored in deontological theory.  There is no reliance on human moral intuitions.  However, (\ref{eq:gen}) is itself an empirical claim that requires human assessment, perhaps with the assistance of mimetic methods.  This suggests a hybrid approach to VA in which both moral anchors and mimesis play a role.

\subsection{Hybrid VA}

Ethical judgment relies on both formal consistency tests and an empirical assessment of beliefs about the world.  Hybrid VA obtains the former using anchored VA and the latter using mimetic VA.  In the example just described, the generalization principle that imposes condition (\ref{eq:gen}) is anchored in ethical analysis, while empirical belief alignment can assess whether (\ref{eq:gen}) is satisfied.  One could take a poll of views on whether I could get away with stealing unguarded merchandise if everyone did so.  Obviously, generally held opinions can be irrational, but they can perhaps serve as a first approximation of rational belief.  

We acknowledge that even {\em ethical} value alignment can play a role in hybrid VA.  However, this is not because because popular moral intuitions should be given epistemic weight, but because they may be part of the fact basis that is relevant to applying anchored ethical principles.  

For example, in some parts of the world, drivers consider it wrong to enter a stream of moving traffic from a side street without waiting for a gap in the traffic.  In other parts of the world this can be acceptable, because drivers in the main thoroughfare expect it and make allowances.   Suppose driver $a$'s action plan is (\ref{eq:action}), where
\[
\begin{array}{l}
C_1(a) = \mbox{Driver $a$ can enter a main thoroughfare by} \\
\hspace{8.5ex} \mbox{moving into the traffic without waiting for}\\
\hspace{8.5ex} \mbox{a gap.} \\
C_2(a) = \mbox{It is considered acceptable in driver $a$'s part} \\
\hspace{8.5ex} \mbox{of the world to move into a stream of traffic} \\
\hspace{8.5ex} \mbox{without waiting for a gap.} \\
A(a) = \mbox{Driver $a$ will move into traffic without waiting} \\
\hspace{7.7ex} \mbox{for a gap.}
\end{array}
\]
The key to checking generalizability of action plan (\ref{eq:action}) is assessing $C_2(a)$.  If $C_2(a)$ is true, moving into the stream of traffic under such conditions is generalizable because it is already generalized, and traffic continues to flow because drivers expect it and have social norms for yielding the right of way in such cases.  

Mimetic VA comes into play because assessing $C_2(a)$ is a matter of reading generally held ethical values.  The goal of doing so, however, is not to align agent $a$'s ethical principles with popular opinion, but to assess an empirical claim about popular values that is relevant to applying the generalization test.

Popular values in the sense of {\em preferences} can also be relevant in a hybrid VA scheme.  This is clearest for the utilitarian principle, which says roughly that an action plan should create no less total net expected utility (welfare) and any other available action plan that satisfies the generalization and autonomy principles.  If people generally prefer a certain state of affairs, one might infer that it creates more utility for them.  The inference is hardly foolproof, but it can be a useful approximation.  Again, there is no presumption that we should agree with popular preferences, but only that the state of popular preferences can be relevant to applying the utilitarian test.

Even the {\em autonomy principle} can benefit from VA.  The principle states roughly that an action plan should not interfere with the ethical action plans of other agents without their informed or implied consent.  Aggressively wedging into a stream of traffic could violate autonomy, unless drivers give implied consent to such interference.  It could even cause injuries that violate the autonomy of passengers as well as the driver. However, if wedging into traffic is standard and accepted practice, perhaps drivers implicitly consent to such interference.  They implicitly agree to play by the rules of the road when they get behind the wheel, whether the rules are legally mandated or a matter of social custom.  The social norms that regulate this kind of traffic flow also help to avoid accidents.   Here again, hybrid VA does not adopt popular values but only acknowledges that they are relevant to applying ethical principles.

Such a hybrid approach is suitable for AI ethics because a machine's instructions are naturally viewed as conditional action plans much like those discussed here.  If deep learning is used, the antecedents of the action plans can include the output of neural networks.  The ethical status of the instructions can then be evaluated by assessing the truth of a generalizability condition similar to (\ref{eq:gen}) for each instruction, where mimetic VA plays a central role in this assessment.  This is analogous to the common practice of applying statistic criteria for fairness \cite{CorbettDavisesGoel2018}, except that the criteria are derived from general ethical principles that are anchored in a logical analysis of action.

\section{Conclusion}
The preceding discussion reveals stumbling blocks for VA approaches that neglect implications of the naturalistic fallacy. Such problems are more serious in \textit{mimetic} VA since the mimetic process imitates human behavior that may or may not rise to the level of correct ethical behavior. Anchored VA, including hybrid VA, in contrast, holds more promise for future VA since it anchors alignment by normative concepts of intrinsic value.

\section*{Acknowledgements}
We would like to express our appreciation to David Danks, Patrick Lin, Zach Lipton, and Bryan Routledge for their valuable and constructive suggestions.

\bibliographystyle{aaai}
\bibliography{VA}

\bigskip

\end{document}